\documentclass[10pt,twocolumn,letterpaper]{article}

\usepackage{cvpr24-template/cvpr}              

\usepackage[dvipsnames]{xcolor}

\usepackage{url}
\usepackage{afterpage}
\usepackage{tabularx}
\usepackage{tikz}
\usepackage{graphicx}
\usepackage{caption}
\usepackage{amsmath}
\usepackage{dblfloatfix}

\definecolor{cvprblue}{rgb}{0.21,0.49,0.74}
\usepackage[pagebackref,breaklinks,colorlinks,citecolor=cvprblue]{hyperref}

\def\paperID{9} 
\def\confName{CVPR}
\def\confYear{2024}

\title{A Survey on 3D Egocentric Human Pose Estimation}

\author{Md Mushfiqur Azam, Kevin Desai\\
The University of Texas at San Antonio\\
{\tt\small \{mdmushfiqur.azam, kevin.desai\}@utsa.edu}
}

\begin{document}
\maketitle

\begin{abstract}
Egocentric human pose estimation aims to estimate human body poses and develop body representations from a first-person camera perspective. It has gained vast popularity in recent years because of its wide range of applications in sectors like XR-technologies, human-computer interaction, and fitness tracking. However, to the best of our knowledge, there is no systematic literature review based on the proposed solutions regarding egocentric 3D human pose estimation. To that end, the aim of this survey paper is to provide an extensive overview of the current state of egocentric pose estimation research. In this paper, we categorize and discuss the popular datasets and the different pose estimation models, highlighting the strengths and weaknesses of different methods by comparative analysis. This survey can be a valuable resource for both researchers and practitioners in the field, offering insights into key concepts and cutting-edge solutions in egocentric pose estimation, its wide-ranging applications, as well as the open problems with future scope.

\end{abstract}   
\section{Introduction}
\label{sec:intro}
\begin{figure}
    \begin{subfigure}{0.98\linewidth}
    \centering
        \includegraphics[height=1.4in]{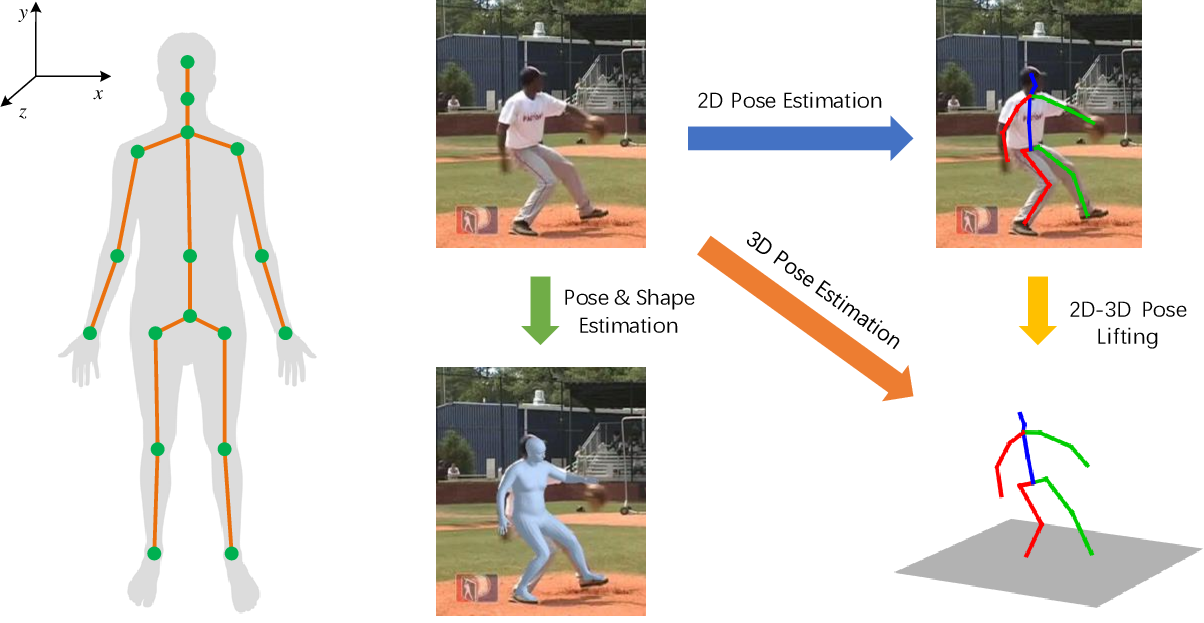}
        \caption{}
        \label{fig:subfig1}
    \end{subfigure}
    \begin{subfigure}{0.999\linewidth}
        \centering
        \includegraphics[width=0.95\linewidth]{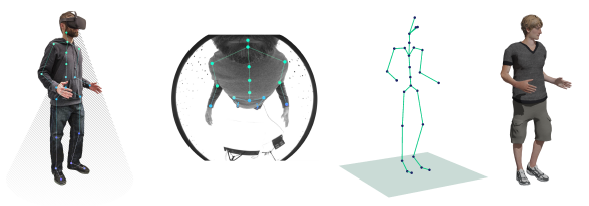}
        \caption{}
        \label{fig:subfig2}
    \end{subfigure}
    \vspace{-7mm}
    \caption{Difference between (a) traditional human pose estimation \cite{ref05} and (b) egocentric human pose estimation \cite{selfpose}}
    \label{fig:pose_estimation}
\vspace{-5mm}
\end{figure}

Human pose estimation \cite{ref01,ref02,ref03,ref04} has gained prominence due to its relevance in numerous applications, ranging from animation and gaming to surveillance, healthcare, and human-computer interaction.
The rise of wearable technology, including smart glasses, body-mounted cameras, and head-mounted displays has significantly fueled interest in egocentric pose estimation, where the focus is on estimating the pose of the person from the point of view of a wearable camera or device worn by the person (first person perspective).
Egocentric pose estimation plays a crucial role across various domains, such as in human computer interaction for gesture recognition, augmented and virtual reality experiences by tracking body movements, healthcare for precise therapy monitoring, biomechanical analysis in sports training, hand-object interaction for contextual understanding, and enhancing realism in professional simulations through accurate movement replication.
Unlike traditional pose estimation, which relies on external cameras or sensors, egocentric pose estimation offers a unique and immersive perspective on human body representation. Real-time processing, adaptability to different environments, user interaction mechanisms, including gestures, and semantic scene understanding contribute to the effectiveness of egocentric pose estimation systems. 
Figure \ref{fig:pose_estimation} shows the difference between traditional and egocentric 3D human pose estimation.

\textbf{\textit{Challenges for Egocentric 3D Human Pose Estimation}} stem from the complexity of accurately capturing and interpreting human movements from the first-person perspective. Some of the key challenges include:
\begin{itemize}
    \item \textit{Viewpoint Variations:} The use of egocentric cameras, attached to the body, introduces challenges in pose estimation as body parts may be occluded, particularly when hidden from view. The wide range of possible viewpoints in egocentric settings, involving varying camera angles, heights, and orientations, demands robust models to ensure accurate pose estimation across diverse scenarios.
    \item \textit{Limited Depth Information:} Egocentric cameras, commonly mounted on wearable devices, capture scenes in 2D, lacking explicit depth details. This absence complicates the accurate determination of the distance of body parts from the camera, as 2D images may project objects at different distances onto the same plane.
    \item \textit{Dataset Constraints:} In-the-wild datasets are essential for capturing real-world complexity, including variations in lighting, backgrounds, activities, and environments. However, their scarcity hinders model generalization, especially in dynamic environments with unpredictable situations. Limited availability of diverse samples, often from motion capture systems, poses challenges for models aiming at real-world outdoor applications.
\end{itemize}

\textbf{\textit{Scope of the Survey:}}
Currently, there are numerous systematic surveys related to 2D and 3D human pose estimation on traditional and deep learning based approaches \cite{ref06,ref07,ref08,ref09} as well shape recovery based approaches \cite{meshsurvey1,meshsurvey2}.
While comprehensive reviews on hand pose \cite{handsurvey} and action recognition \cite{actionsurveyegocentric} from egocentric vision are present, it is noteworthy that, to the best of our knowledge, no comprehensive survey on full body egocentric 3D pose estimation methods has been published to date. This absence underscores a notable gap in existing research, despite the increasing interest and advancement in this domain.

In this survey, we aim to explore the multifaceted aspects of 3D egocentric human pose estimation, by first describing the widely used datasets in Section \ref{sec:datasets}.
Next, in Section \ref{sec:7_egocentric_pose_estimation}, we explore the different egocentric estimation methods by dividing them into two categories on the basis of output generation: skeletal based methods and human body shape based methods. Skeletal methods explore different methods which are mostly regression based (estimation of 3D joint co-ordinates) and heatmap based (estimation of 2D heatmaps). On the other hand, body shape based methods mainly generate human models using different shape recovery methods.
Additionally, we present a comprehensive evaluation of egocentric pose estimation models, showcasing various evaluation metrics in Section \ref{sec:6_evaluation_metrics} and a detailed performance analysis of state-of-the-art approaches on prominent datasets in Section \ref{sec:8_performance_analysis}.
Lastly, we conclude the survey in Section \ref{sec:9_conclusion_future_direction} with some future research scopes for egocentric 3D human pose estimation.

\section{Datasets}
\label{sec:datasets}
\begin{figure}[!b]
 \vspace{-5mm}
  \begin{minipage}[b]{0.99\linewidth}
    \centering
    \includegraphics[height=30mm]{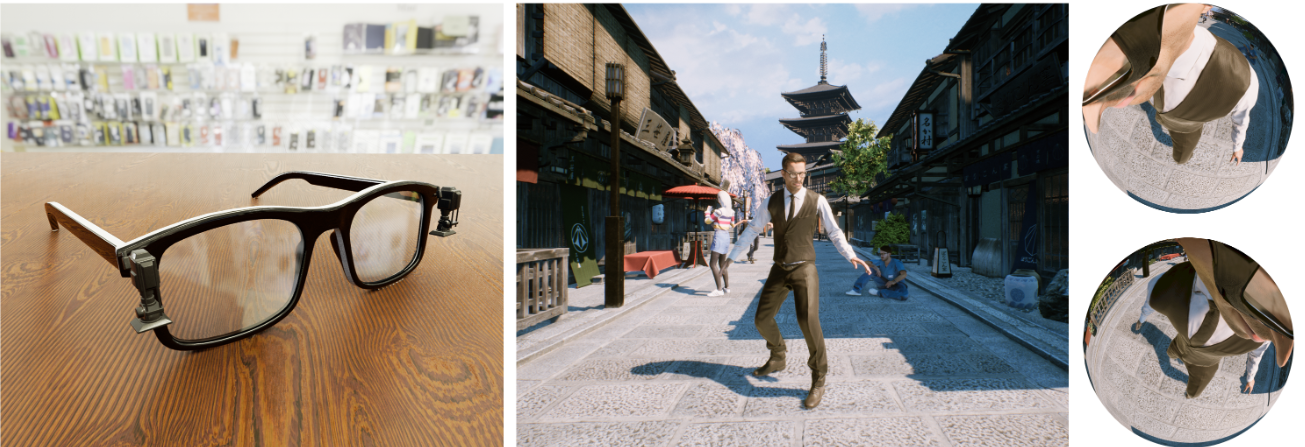}
    \captionsetup{justification=centering}
    \subcaption{Dataset setup for UnrealEgo \cite{unrealego}: Left image shows a glass equipped with two fisheye cameras. The middle image provides a third-person perspective of the person, offering context to the scene. The right image depicts the egocentric view of the person.}
    \label{fig:data_subfig1}
  \end{minipage}

  \begin{minipage}[b]{\columnwidth}
    \centering
    \captionsetup{justification=centering}
    \includegraphics[height=25mm, width=30mm]{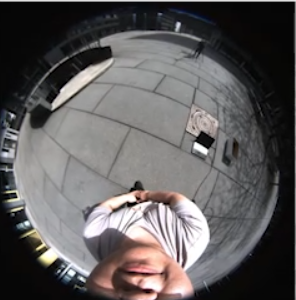}
    \includegraphics[height=25mm,width =30mm]{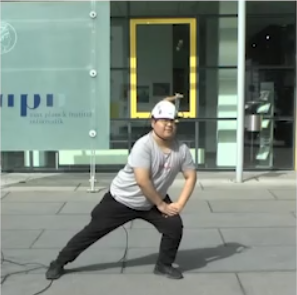}
    \subcaption{Sample image from EgoPW \cite{egopw} dataset visualizing egocentric view on the left image and exocentric view on the right image.}
    \label{fig:data_subfig3}
  \end{minipage}\hfill
  \begin{minipage}[b]{0.5\columnwidth}
    \centering
     \captionsetup{justification=centering}
    \includegraphics[height=25mm]{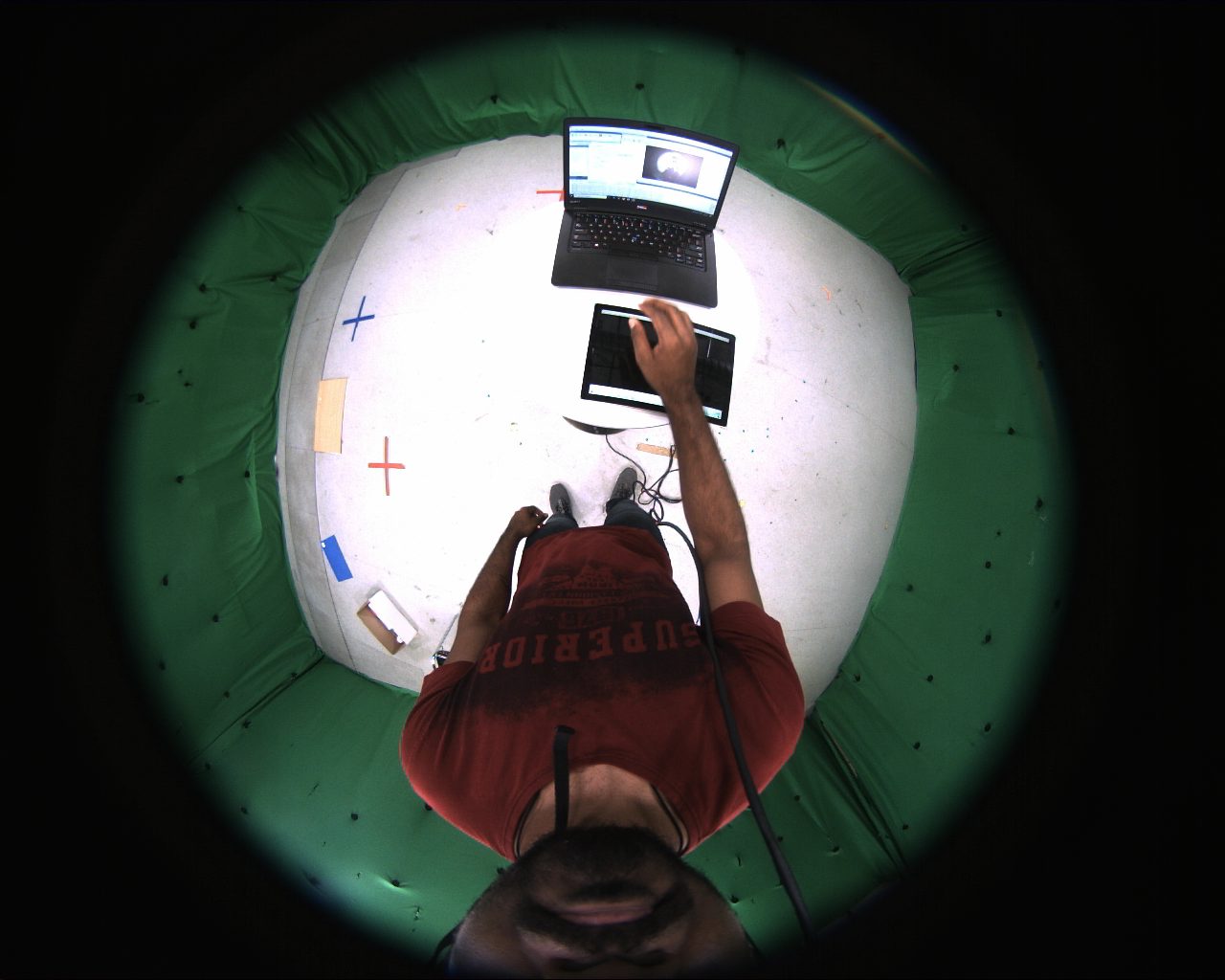}
    \subcaption{ Sample image from EgoGTA \cite{sceneaware} dataset.}
    \label{fig:data_subfig4}
  \end{minipage}\hfill
  \begin{minipage}[b]{0.5\columnwidth}
    \centering
    \includegraphics[height=25mm]{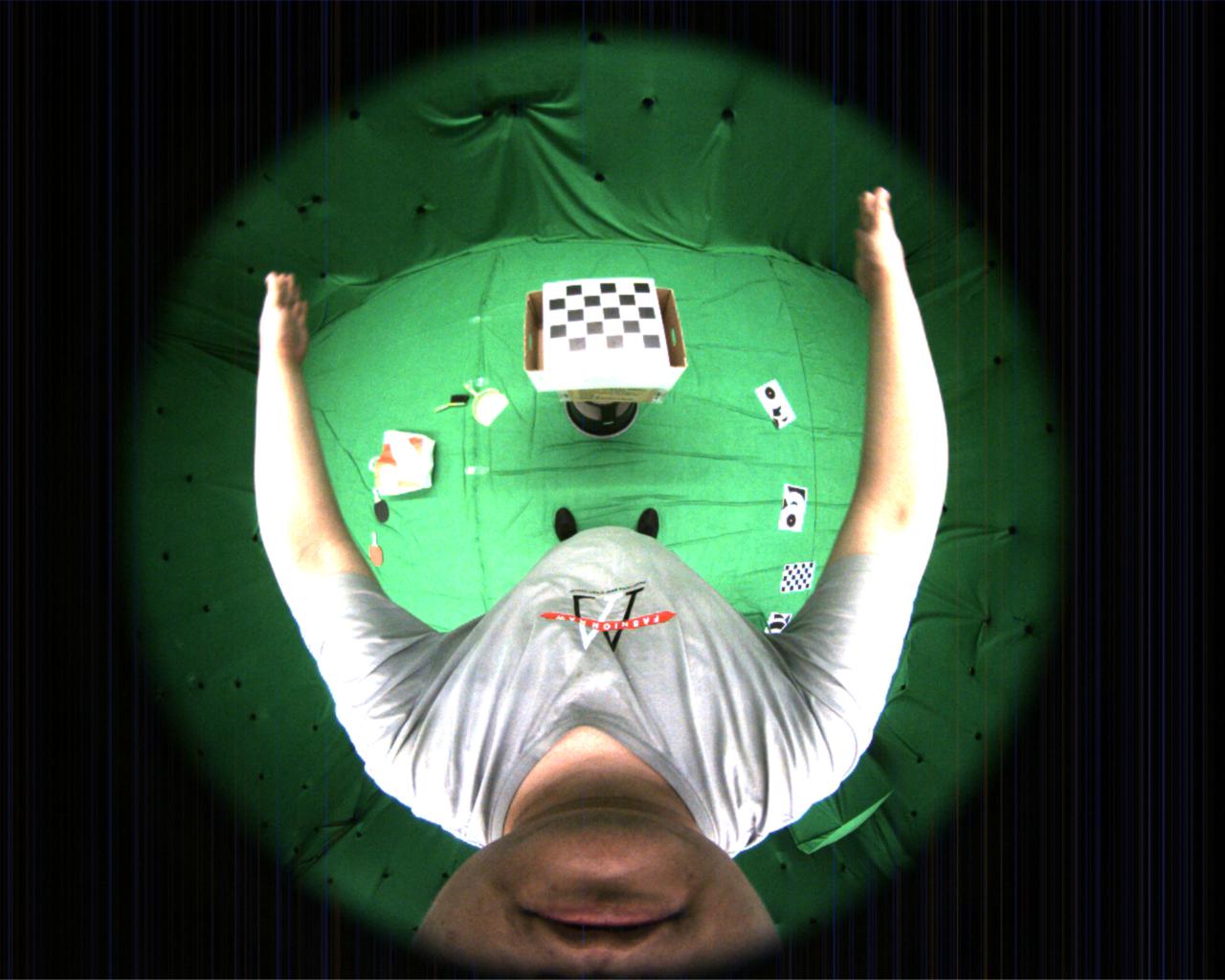}
    \subcaption{Sample image from Wang et al.'s \cite{egoglobal} dataset.}
    \label{fig:data_subfig5}
  \end{minipage}
  \caption{Sample images from different datasets used for egocentric human pose estimation.}
  \label{fig:dataset}
\end{figure}

Large scale dataset is one of the key factors in visualizing and  analysing a computer vision problem. While benchmark datasets like MPII \cite{mpii} and Human3.6M \cite{human3.6m} exist for traditional human pose estimation, there's a notable gap for egocentric pose estimation benchmark datasets. Figure \ref{fig:dataset} showcases sample images from 4 different datasets. Table \ref{tab:dataset} summarizes the key features of 9 egocentric pose estimation datasets, with more details provided in the text below.

\textbf{\textit{EgoCap}} \cite{egocap} proposed a method for creating large training datasets using a marker-less motion capture system. They leveraged eight fixed cameras to estimate 3D skeleton motion. They projected it onto fisheye images from a head-mounted camera setup, enhancing the dataset with background replacement, clothing color variations, and simulated lighting changes.
The training set includes 75,000 annotated fisheye images from six subjects and 25,000 images from two additional subjects for validation.

\begin{table*}[ht]
    \scriptsize
    \centering
    \begin{tabularx}{\textwidth}{|>{\centering\arraybackslash}p{1.7cm}|>{\centering\arraybackslash}p{6mm}|>{\centering\arraybackslash}p{2.25 cm}|>{\centering\arraybackslash}p{2.7 cm}|>
    {\centering\arraybackslash}p{5.9 cm}|>
    {\centering\arraybackslash} X|}
        \hline
        \textbf{Dataset} & \textbf{Year} & \textbf{No. of Images} & \textbf{No. of Subjects / Actions} & \textbf{Characteristics} & \textbf{Dataset Website} \\
        \hline
        EgoCap \cite{egocap} & 2016 & 100,000 & 8 subjects & marker-less motion capture system; annotated. & \href{https://vcai.mpi-inf.mpg.de/projects/EgoCap/}{Link} \\ \hline
        Mo\textsuperscript{2}Cap\textsuperscript{2}\cite{mo2cap2} & 2019 & 530,000 & 3000 actions & annotated; 700 different body textures. & \href{https://vcai.mpi-inf.mpg.de/projects/wxu/Mo2Cap2/}{Link}\\ \hline
        xr-EgoPose \cite{xregopose} & 2019 & 383,000 & 23 male and 23 female subjects; 9 actions & synthetic; scene is generated from randomized characters, environments, lighting rigs and animation. & \href{https://github.com/facebookresearch/xR-EgoPose/tree/main}{Link} \\ \hline
        EgoBody \cite{egobody} & 2022 & 219,731 & 15 indoor scenes; 36 subjects & two subjects (camera wearer and interactee) involved in different interaction scenarios. & \href{https://sanweiliti.github.io/egobody/egobody.html}{Link}\\ \hline
        EgoPW \cite{egopw} & 2022 & 318,000 & 10 subjects; 20 actions & in-the-wild real data; 20 different clothing styles. & \href{https://people.mpi-inf.mpg.de/~jianwang/projects/egopw/}{Link}\\ \hline
        UnrealEgo \cite{unrealego} & 2022 &  900,000 & 17 subjects; 30 actions  & 450k in-the-wild stereo views; Motions, 3D environments, spawning human characters. & \href{https://4dqv.mpi-inf.mpg.de/UnrealEgo/}{Link} \\ \hline
        EgoGTA \cite{sceneaware} & 2023 & 320,000 & 101 different actions & synthetic; based on GTA-IM containing different daily motions and scene geometry. & \href{https://people.mpi-inf.mpg.de/~jianwang/projects/sceneego/}{Link} \\ \hline
        ECHP \cite{egofish3d} & 2023 & 30 video sequences; 75,000 frames & 9 subjects; 10 daily actions & indoor and outdoor; real-world data. & \href{https://github.com/Lrnyux/EgoCentric-Human-Pose-ECHP-Dataset}{Link} \\ \hline
        Ego-Exo4D \cite{egoexo} & 2023 & 5625 video sequences; 1422 hours & 839 subjects; 43 actions & 131 different scenes in 13 different cities; comprises skilled human activities (e.g., sports, music, dance, bike repair). & \href{https://docs.ego-exo4d-data.org/}{Link}\\
        \hline
    \end{tabularx}
  \vspace{-2mm}
  \caption{Popular datasets for egocentric 3D human pose estimation.}
  \label{tab:dataset}
\vspace{-5mm}
\end{table*}

The \textbf{\textit{Mo\textsuperscript{2}Cap\textsuperscript{2}}} dataset \cite{mo2cap2} tackles the challenge of obtaining annotated 3D pose data and introduces a marker-less multi-view motion capture. To address the time-consuming nature of obtaining diverse egocentric training examples, the dataset includes a synthetic training corpus generated from egocentric fisheye views. Built upon the SURREAL dataset \cite{surreal}, it offers 530,000 realistic training images with ground truth annotations of 2D and 3D joint positions.

\textbf{\textit{xr-EgoPose}} \cite{xregopose} provides an extensive collection of 383,000 frames featuring individuals showcasing a rich diversity of skin tones, body shapes, clothing styles, set with various backgrounds and lighting scenarios. Scenes are randomly generated from mocap data, featuring realistic body types like skinny short to full tall versions and skin tones from white to black. Prioritizing photorealism, the synthetic dataset is created through Maya animation with mocap data and V-Ray's physically based rendering setup.


\textbf{\textit{EgoBody}} \cite{egobody} captures 2-person interactions using a Microsoft HoloLens2 headset. It provides synchronized multi-modal data, including RGB, depth, head, hand, and eye gaze tracking. With 125 sequences from 36 subjects in 15 scenes, it offers accurate 3D human shape, pose, and motion ground-truth. The dataset aims to explore the relationships between human attention, interactions, and motions, overcoming limitations of prior datasets, and advancing sociological and human-computer interaction research.

The \textbf{\textit{EgoPW}} \cite{egopw} dataset is the first in-the-wild human performance dataset captured by synchronized egocentric and external cameras. It features 10 actors, 20 clothing styles, and 20 actions from 318,000 frames organized into 97 sequences, along with the 3D poses as pseudo labels.

The \textbf{\textit{UnrealEgo}} dataset \cite{unrealego} introduces robust egocentric 3D human motion capture with 17 diverse 3D models and over 45,000 motions in 14 environments. It has stereo fisheye images and depth maps capturing complex activities like breakdance and backflips. With metadata including 3D joint positions and camera details, it comprises 450,000 in-the-wild stereo views, showcasing wider joint position distributions compared to xR-EgoPose \cite{xregopose}. They followed up with the \textbf{UnrealEgo2} and \textbf{UnrealEgo-RW} datasets \cite{unrealego2}, which provide more views with diverse human motions.

The \textbf{\textit{EgoGTA}} \cite{sceneaware} dataset comprises of 320,000 frames across 101 sequences with distinct human body textures, by leveraging the diverse daily motions and ground truth scene geometry of GTA-IM \cite{scenecontext}. The methodology involves fitting the SMPL-X \cite{smplx} model to 3D joint trajectories from GTA-IM \cite{scenecontext}, followed by attaching a virtual fisheye camera to the forehead for generating synthetic images, semantic labels, and depth maps with and without the human body.

The \textbf{\textit{ECHP}} \cite{egofish3d} dataset consists of 65,000 training images, 10,000 validation images, and a test set with egocentric images and 3D ground truth from VICON Mocap. Egocentric poses are extracted using OpenPose \cite{ref02} and human segmentation. Calibration and Aruco markers \cite{arucomarker} aid in obtaining egocentric camera pose. The dataset has 30 sequences with 9 subjects, 20 textures, and 10 actions in various indoor/outdoor scenes. The test set provides generalization with 4 unseen subjects and 17,000 ground truth frames.

 
\textbf{\textit{Ego-Exo4D}} \cite{egoexo} is a groundbreaking multimodal dataset and benchmark suite, offering the largest public collection of time-synchronized first and third-person videos captured by 839 individuals across 131 scenes in 13 cities. It is comprised of 1,422 hours of video, featuring both egocentric and multiple synchronized exocentric views. The \textbf{EgoPose} benchmark focuses on recovering 3D body and hand movements from egocentric videos. The task is to estimate 17 3D body joint positions and 21 3D joint positions per hand, following the MS COCO convention.

\begin{table*}[ht]
\scriptsize
    \centering
     \resizebox{\textwidth}{!}{
      \begin{tabularx}{\textwidth }{|>{\centering\arraybackslash}p{1.8cm}|>{\centering\arraybackslash}p{5mm}|>{\centering\arraybackslash}p{6.3cm}|>{\centering\arraybackslash}p{1.45cm}|>{\centering\arraybackslash}p{3.3cm}|>
      {\centering\arraybackslash} X|}
        \hline
        \textbf{Skeletal based Methods} & \textbf{Year} & \textbf{Highlighted Characteristics} & \textbf{Dataset} & \textbf{Limitations} & \textbf{Code/Project Website Link} \\
        \hline 
        Egocap \cite{egocap} & 2016 & First marker-less motion capture system; utilized pose estimation framework for fisheye views with a ConvNet based body-part detector.& EgoCap & No real-time prototype. & \href{https://vcai.mpi-inf.mpg.de/projects/EgoCap/}{Project} \\\hline 
        Jiang et al. \cite{invisible} & 2017 & Leveraged dynamic motion signatures and static scene structures to infer the invisible pose efficiently. & custom Kinect V2 dataset & Ambiguity in egocentric inputs due to unpredictable arm poses. & --\\\hline 
        Mo\textsuperscript{2}Cap\textsuperscript{2} \cite{mo2cap2} & 2019 & Real-time; disentangled 3D pose estimation, addressed 2D joint detection, camera-to-joint distances, and joint position recovery for accurate results and a precise 2D overlay. & Mo\textsuperscript{2}Cap\textsuperscript{2} & Scenes with severe occlusions. & \href{https://vcai.mpi-inf.mpg.de/projects/wxu/Mo2Cap2/}{Project} \\\hline 
        xr-EgoPose \cite{xregopose} & 2019 & Encoder-decoder model for VR headset images, addressing resolution differences in upper and lower body poses, with a dual-branch decoder preserving uncertainty information. & xR-EgoPose & Scenes with extreme occlusions and out-of-field view. & \href{https://github.com/facebookresearch/xR-EgoPose}{Code} \\ \hline
        You2Me \cite{you2me} & 2020 & Inferred robust poses by incorporating static scene features, explicit second-person body interactions and utilizing dyadic interactions and dynamic first-person motion features. & You2Me & Scenerios where camera wearer is crouched and camera points towards the floor. & \href{https://github.com/facebookresearch/you2me}{Code} \\ \hline
        EgoGlass \cite{egoglass} & 2021 & Utilized body part information for low-visible joints and tackling self-occlusion by preserving uncertainty information. & EgoGlass & Lower body estimation produces larger errors. & -- \\\hline  
        Zhang et al. \cite{autocalib} & 2021 & Implemented auto-calibration module with self-correction for fisheye cameras to rectify image distortions, ensuring alignment between 3D predictions and distorted 2D poses. & xR-EgoPose & Not evaluated in real-world setting. & -- \\ \hline
        Wang et al. \cite{egoglobal} & 2021 & Spatio-temporal optimization framework that combines 2D and 3D keypoints, VAE-based motion priors and SLAM-based camera pose estimation for stable global body pose estimation. & Mo\textsuperscript{2}Cap\textsuperscript{2}, AMASS & Not evaluated in real-world setting. & \href{https://people.mpi-inf.mpg.de/~jianwang/projects/globalegomocap/}{Project} \\ \hline
        Wang et al. \cite{egopw} & 2022 & Implemented weak supervision with spatio-temporal optimization and synthetic data with domain adaptation for better egocentric pose estimation. & EgoPW & Accuracy of pseudo labels constrained by in-the-wild capture system. & -- \\ \hline 
        Akada et al. \cite{unrealego} & 2022 & Enhances 3D pose estimation by integrating a stereo-based 2D joint location estimation module with weight-sharing encoders and a multi-branch autoencoder for uncertainty capture. & UnrealEgo & Occlusions and complex motions scenerios. & \href{https://4dqv.mpi-inf.mpg.de/UnrealEgo/}{Project}\\ \hline
        Ego+X \cite{ego+x} & 2022 & Dual-camera framework for 3D global pose estimation and social interaction characterization, leveraging visual SLAM and a Pose Refine Module (PRM) for spatial and temporal accuracy and characterizes social interactions based on global 3D poses. & ECHA & Camera localization robustness limited; temporal smoothing effectiveness not fully evaluated. & --\\ \hline
        Wang et al. \cite{sceneaware} & 2023 & First egocentric pose estimation framework, integrating depth estimation for occlusion handling in close interactions. & EgoGTA, EgoPW-Scene & Accuracy is constrained by depth estimation where scene is occluded by body. & \href{https://people.mpi-inf.mpg.de/~jianwang/projects/sceneego/}{Project}\\ \hline
        EgoFish3D \cite{egofish3d} & 2023 & A self-supervised framework for egocentric 3D pose estimation, utilizing real-world data with three key modules: third person view, egocentric, and interactive modules, achieving accurate results without the need for ground truth annotations. & ECHP & Overlooked the significance of the perspective factor, which can convey valuable information about the 3D effect intensity. & -- \\ \hline
        Ego3DPose \cite{ego3dpose} & 2023 & A stereo matcher network and perspective embedding heatmap representation, independent learning of stereo correspondences and leveraging 3D perspective information. & UnrealEgo & Scenes with occlusions, distortions and real-world setting. & \href{https://github.com/tho-kn/Ego3DPose}{Code} \\ \hline
        Ego-STAN \cite{egostan} & 2023 & Tackles fisheye distortion and self-occlusions in egocentric human pose estimation through a domain-guided spatio-temporal transformer, using 2D image representations, feature map tokens, and 3D pose estimation for accurate joint localization and uncertainty management.& xr-EgoPose & Scenes in real-world setting. & --\\ \hline
        Dhamanaskar et al. \cite{first2third} & 2023 & Utilized third-person view information, creating a self-supervised neural network that establishes a shared space for consistent 3D body pose detection across diverse video settings, ensuring adaptability to real-world scenarios with unknown camera configurations. & First2Third-Pose & Evaluation limited to two datasets; broader assessment needed for generalization. & \href{https://github.com/nudlesoup/First2Third-Pose}{Code}\\ \hline
        EgoFormer \cite{egoformer} & 2023 & Leveraged video context and establishing long-term temporal relationships. It addresses ambiguity in first-person videos, surpassing dynamic features, and introduces a novel motion clue representation for enhanced accuracy. & CMU Mocap \cite{cmumocap} & Lack of real-world testing and limited model comparisons. & -- \\ \hline
        
      \end{tabularx}
      }
  \vspace{-2mm}
  \caption{Popular skeletal based egocentric 3D pose estimation methods.}
  \label{tab:skeletal}
\vspace{-4mm}
\end{table*} 

\section{3D Egocentric Pose Estimation Methods}
\label{sec:7_egocentric_pose_estimation}
After performing an extensive literature search for egocentric pose estimation, in this section, we discuss around 35 popular techniques by classifying them into two categories, namely skeletal and body shape based approaches.
Skeletal-based 3D pose estimation methods \cite{skeletal1,skeletal2,skeletal3} leverage the human skeleton representation to accurately track and infer 3D joint position and body movements.
Human body shape based human body pose estimation methods \cite{keepitsmpl, monocularshape,towardsaccurate} utilize a parametric model, such as SMPL \cite{smpl} and SMPL-X \cite{smplx}, to accurately estimate 3D joint locations and body shapes. The retrieval of human body meshes is pivotal in supporting subsequent tasks like reconstructing clothed humans \cite{clothed1,clothed2}, rendering \cite{rendering}, and modeling avatars \cite{arch,avatarrex}.
The sub-sections below expand on the different methods in each category.
We have further sub-categorized the methods based on some significant features, as highlighted in bold.

\subsection{Skeletal Based Methods}
In this section, we have provided details on the skeletal based egocentric pose estimation methods. Table \ref{tab:skeletal} provides a brief overview of 17 such skeletal based methods.

Rhodin et al. \cite{egocap} introduced a \textbf{marker-less} egocentric motion capture system using fisheye cameras embedded in a helmet or VR headset. The method employs a generative pose estimation framework with a ConvNet-based body part detector, ideal for VR applications needing natural movement and interaction. However, it was not able to attain \textbf{real-time} performance.
To solve which, Mo\textsuperscript{2}Cap\textsuperscript{2} \cite{mo2cap2} uses a two-scale location invariant convolutional network to detect 2D joints, accommodating perspective and radial distortions. It uses a location-sensitive distance module for estimating absolute camera-to-joint distances, and then recovers actual joint positions by back-projecting 2D detections. However, it struggles in scenes with severe occlusions. 

\textit{EgoGlass} \cite{egoglass} solves the \textbf{occlusion} problem by leveraging body part information for improved pose detection. The 2D module incorporates branches for heatmap and body part prediction, while the 3D module employs a pseudo-limb mask approach to handle occlusion in real-world images. This module also functions as an autoencoder for joint heatmaps, enhancing 3D body pose estimation and capturing uncertainty in 2D predictions across multiple views.
\cite{sceneaware} introduces an egocentric depth estimation network for predicting scene depth maps behind the human body using a wide-view egocentric fisheye camera, addressing occlusion caused by the human body through a depth-inpainting network. Additionally, a scene-aware pose estimation network was presented for 3D pose regression.
\cite{doublediscrete} used a Vector Quantized-Variational AutoEncoder (VQ-VAE) to predict and optimize human pose, addressing the challenge of obscured lower body appearance.
\textit{xR-EgoPose} \cite{xregopose} and \textit{SelfPose} \cite{selfpose} uses an encoder-decoder architecture designed to improve accuracy in capturing upper and lower body poses from monocular images obtained via VR headset cameras. \cite{xregopose} employs a dual-branch decoder to address resolution discrepancies between the upper and lower body. It handles uncertainties in 2D joint locations by initially generating 2D heatmaps and subsequently using an autoencoder for 3D pose regression.

To solve the problem of \textbf{out-of-field-view}, \cite{invisible} developed a method aiming to infer the invisible pose of a person in egocentric videos using dynamic motion signatures and static scene structures. By combining short-term and longer-term pose dynamics, the method utilizes classifiers to estimate pose probabilities and performs joint inference for a longer sequence. They extended the idea \cite{visionspan} by using both dynamic motion information from camera SLAM and occasionally visible body parts for robust ego pose estimation ensuring geometrical consistency.
\textit{EgoTAP} \cite{egotap} addresses \textbf{out-of-view limbs} and self-occlusion issues in stereo egocentric 3D pose estimation by introducing a Grid ViT Heatmap Encoder and Propagation Network. The Grid ViT efficiently summarizes joint heatmaps, preserving spatial relationships. The Propagation Network utilizes skeletal information to predict 3D poses, improving accuracy for both visible and less visible joints.


To reduce the scarcity of \textbf{real-world datasets} from egocentric view, \cite{egopw} proposed the use of weak supervision from an external viewpoint. The approach utilizes spatio-temporal optimization to generate accurate 3D poses for frames in the \textit{EgoPW} dataset, using them as labels for training an egocentric pose estimation network. 
It also incorporates a synthetic dataset and employs domain adaptation to bridge the gap between synthetic and real data.
\cite{unrealego} proposed a solution for egocentric pose estimation in an \textbf{unconstrained environment}. It uses a 2D joint location estimation module for stereo inputs by utilizing weight-sharing encoders and a decoder leveraging stereo information to boost performance. The 3D module comprises a multi-branch autoencoder, predicting 2D heatmaps to generate 3D pose and reconstructing heatmaps to capture uncertainty.

\textbf{Perspective distortion} can cause issues like scale variation, depth ambiguity and limited field of view. To tackle this problem, \textit{Ego3DPose} \cite{ego3dpose} introduces a Stereo Matcher network that independently learns stereo correspondences and predicts explicit 3D orientation for each limb, avoiding dependence on full-body information. Additionally, a Perspective Embedding Heatmap representation is introduced, allowing the 2D module to extract and utilize 3D perspective information. \cite{egostan} addressed the challenges of \textbf{fisheye distortion} and \textbf{self-occlusions} by leveraging a domain-guided spatio-temporal transformer model, \textit{Ego-STAN}. It utilizes 2D image representations and spatiotemporal attention to mitigate distortions and accurately estimate the location of heavily occluded joints. 
\cite{autocalib} employed an \textbf{automatic calibration} module with self-correction to mitigate the impact of image distortions on 3D pose estimation. Unlike traditional post-processing steps, this module ensures consistency between 3D predictions and distorted 2D poses.

When the predicted poses are in the fisheye camera's \textbf{local coordinate system} instead of the global coordinate system, it can cause issues like \textbf{temporal instability}. To solve this issue, \cite{egoglobal} proposed a method for precise and stable global body pose estimation in egocentric videos. It utilizes CNN-detected 2D and 3D keypoints, VAE-based motion priors, and SLAM-based camera pose estimation. This approach effectively tackles challenges like temporal jitters and tracking failures, significantly enhancing accuracy and stability in obtaining coherent body poses. \textit{Ego+X} \cite{ego+x} proposed a framework with two cameras for 3D \textbf{global pose estimation} and \textbf{social interaction characterization}. The \textit{Ego-Glo} module solves spatial and temporal errors using a dual-branch network and visual SLAM. Whereas, the \textit{Ego-Soc} module performs egocentric social interaction characterization, including object detection and human-human interaction, based on the global 3D human poses.

\begin{table*}[h]
\scriptsize
    \centering
     \resizebox{\textwidth}{!}{
      \begin{tabularx}{\textwidth }{|>{\centering\arraybackslash}p{1.85cm}|>{\centering\arraybackslash}p{5mm}|>{\centering\arraybackslash}p{6.5cm}|>{\centering\arraybackslash}p{1.15cm}|>{\centering\arraybackslash}p{3.4cm}|>
      {\centering\arraybackslash} X|}
        \hline
        \textbf{Body Shape based Methods} & \textbf{Year} & \textbf{Highlighted Characteristics} & \textbf{Dataset} & \textbf{Limitations} & \textbf{Code/Project Website Link} \\
        \hline 
        Yuan et al. \cite{imitation} & 2018 & Integrates control-based modeling, physics simulation, and imitation learning for ego-pose estimation, enabling domain adaptation by considering underlying physics dynamics. & CMU Mocap \cite{cmumocap} &  Indirect 2D evaluation may not capture full 3D accuracy; limited behaviors may hinder complex motion generalization. & -- \\ \hline 
         Dittadi et al. \cite{partial} & 2021 & Variational autoencoders for generating human body poses from limited head and hand pose data; addressing challenges through specialized inference models. & AMASS \cite{amass} & Incomplete utilization of temporal history, constraints on body shape variation and reliance on assumed availability of hand signals. & --\\ \hline
        CoolMoves \cite{coolmoves} & 2021  & Achieves real-time, expressive full-body motion synthesis for avatars using limited input cues, dynamically fusing stylized examples from skilled performers, excelling in activities like dancing and fighting. & CMU MoCap & Limited sensing of legs and feet, resulting in lower body reconstruction jitters and reduced accuracy in foot-driven motions. & --\\ \hline
        EgoRenderer \cite{egorenderer} & 2021 & Renders full-body neural avatars from egocentric fisheye images - texture synthesis, pose construction, and neural image translation; addresses challenges of top-down view and distortions. & EgoRenderer & Incomplete joint estimation; unnatural motions in SMPL model animations and temporal instability in frame predictions. & \href{https://vcai.mpi-inf.mpg.de/projects/EgoRenderer/}{Project} \\ \hline
        HPS \cite{hps} & 2021 & Integrates wearable sensors, IMUs and a head-mounted camera for precise 3D pose tracking in pre-scanned environments, eliminating drift with localization and scene constraints. & HPS &  Lack of features and scene changes between static 3D scans and real images. & \href{https://virtualhumans.mpi-inf.mpg.de/hps/}{Project} \\ \hline
        Avatarposer \cite{avatarposer} & 2022 & First learning-based method predicting full-body poses in world coordinates, leveraging transformer encoder and motion input from head and hands & CMU Mocap, AMASS &  Sensitivity to inaccuracies and occlusions in hand tracking data. & \href{https://github.com/eth-siplab/AvatarPoser}{Code}\\\hline 
        FLAG \cite{flag} & 2022 & Flow-based model for realistic 3D human body pose prediction with uncertainty estimates, enhancing prior work through high-quality pose generation and efficient latent variable sampling for optimization. & AMASS & Difficulty in generating complex lower-body poses due to sparse training data and lack of temporal data integration. & \href{https://microsoft.github.io/flag/}{Project}\\ \hline
        Su et al. \cite{proprioception} & 2022 & A data framework transforms raw video into 3D pose, enriched by a lightweight Self-Perception Excitation (SPE) module for egocentric self-awareness. & Mocap dataset \cite{forecast} &  Dependency on MoCap data and synchronized third-person view videos may limit the method's real-world applicability. & --\\ \hline
        EgoEgo \cite{egoego} & 2023 &  Ego body pose estimation using ego head pose estimation leveraging SLAM, and conditional diffusion for disentangled head and body pose estimation. & ARES & Evaluation on synthetic and relatively small real-world datasets. & \href{https://github.com/lijiaman/egoego_release}{Code}\\ \hline
        EgoHMR \cite{egohmr} & 2023 & Scene-conditioned diffusion approach using a physics-based collision score, realistic human-scene interaction, accurate estimation for visible body parts while enhancing diversity. & EgoBody & Limited temporal context for reconstructing egocentric human motions. & \href{https://github.com/sanweiliti/EgoHMR}{Code}\\ \hline
        EgoPoser \cite{egoposer} & 2023 & Used sparse motion sensor; mitigates overfitting with position-invariant prediction, adaptable to diverse body sizes, robust with hands out of view and reduces motion artifacts. & AMASS & Limited evaluation on diverse real-world scenarios. & --\\ \hline
        SimpleEgo \cite{simpleego} & 2024 & Directly predicts joint rotations as matrix Fisher distributions, providing robust uncertainty estimation and realistic deployment prospects. & SynthEgo & Accuracy could be limited when large portions of the body are occluded in the image. & \href{https://microsoft.github.io/SimpleEgo/}{Project}\\
        \hline

      \end{tabularx}
      }
  \vspace{-2mm}
  \caption{Popular body shape based egocentric 3D pose estimation models.}
  \label{tab:shaperecovery}
\vspace{-5mm}
\end{table*}

Generating \textbf{3D ground truth} data using motion capture system is a cumbersome task. To alleviate this problem, \textit{EgoFish3D} \cite{egofish3d} proposed three modules: a third-person view module generating accurate 3D poses from external camera images, an egocentric module predicting 3D poses from a single fisheye image via self-supervised learning, and an interactive module estimating rotation differences between third-person and egocentric views. This method achieves self-supervised egocentric 3D pose estimation without ground truth annotations, leveraging a real-world dataset (ECHP) with synchronized third-person and egocentric images.
\textbf{Linking first-person and third-person view} \cite{ego-exo, third-person1, third-person2} plays a crucial role for better understanding wearer's action and poses. \cite{dhamanaskar} used visual information from paired third-person videos to create a shared space where different views of the same pose are close together. They trained a special neural network to learn this shared space in a self-supervised manner, teaching it to distinguish if two views show the same 3D skeleton.

\textit{EgoFormer} \cite{egoformer}, a tansformer-based model for ego-pose estimation in AR and VR applications, addresses the ambiguity in first-person videos by leveraging video context and establishing long-term temporal relationships. It extracts effective temporal features, dynamic features, and introduces a novel representation for motion clues.
\textit{You2Me} \cite{you2me} addresses the challenge of estimating the 3D body pose of the camera wearer by leveraging interactions with a \textbf{visible second person}. The key insight is that dyadic interactions between individuals help to learn temporal models for interlinked poses even when one person is largely \textbf{out of the field view}. The method incorporates dynamic first-person motion features, static first-person scene features, and second-person's body pose interaction features to explicitly account for the body pose of the camera wearer.

\subsection{Body Shape Based Methods} 
In this section, we expand on the different human body shape-based egocentric pose estimation methods found in the literature. Out of them, 12 are highlighted in Table \ref{tab:shaperecovery}.

  
  
  

\begin{table*}[!b]
    \vspace{-3mm}
    \centering
    \small
    \begin{tabularx}{\textwidth}{|>{\centering\arraybackslash}p{2.75cm}|>{\centering\arraybackslash}X|>
    {\centering\arraybackslash}X|>{\centering\arraybackslash}p{1.5cm}|>{\centering\arraybackslash}p{1.5cm}|>
    {\centering\arraybackslash}X|>
    {\centering\arraybackslash}X|>
    {\centering\arraybackslash}p{1.5cm}|>
    {\centering\arraybackslash}X|>
    {\centering\arraybackslash} X|}
        \hline 
        \textbf{Methods} & \textbf{Walking} & \textbf{Sitting} & \textbf{Crawling} & \textbf{Crouching} & \textbf{Boxing} & \textbf{Dancing} & \textbf{Stretching} & \textbf{Waving} & 
        \textbf{Average}\\ \hline
    EgoFish3D \cite{egofish3d} & 60.9 & 42.1 & 65.0 & 82.7 & 79.0 & 55.5 & 59.1 & 94.5 & 66.8\\ \hline
    Zhang et al. \cite{autocalib} & 41.16 & 76.58 & 73.04 & 89.67 & 52.96 & 58.90 & 92.21 & 71.55 & 62.13 \\ \hline
    Mo\textsuperscript{2}Cap\textsuperscript{2} \cite{mo2cap2} & 38.41 & 70.94 & 94.31 & 81.90 & 48.55 & 55.19 & 99.34 & 60.92 & 61.40 \\ \hline
    xR-EgoPose \cite{xregopose} & 38.39 & 61.59 & 69.53 & 51.14 & 37.67 & 42.10 & 58.32 & 44.77 & 48.16 \\ \hline
 
    \textbf{SelfPose-UNet} \cite{selfpose} & 45.83 & 47.24 & 47.35 & 45.15 & 48.72 & 47.00 & 46.15 & 46.45 & \textbf{46.61}\\ \hline

    \end{tabularx}
  \vspace{-3mm}
  \caption{Comparison of different skeletal based 3D egocentric pose estimation methods on Mo\textsuperscript{2}Cap\textsuperscript{2} dataset using  MPJPE (mm).}
  \label{tab:mo2cap2}
\end{table*}

\begin{table*}[!b]
\vspace{-2mm}
    \centering
    \small
    \begin{tabularx}{\textwidth}{|>{\centering\arraybackslash}p{2.75cm}|>{\centering\arraybackslash}X|>
    {\centering\arraybackslash}X|>{\centering\arraybackslash}p{1.5cm}|>{\centering\arraybackslash}X|>
    {\centering\arraybackslash}X|>
    {\centering\arraybackslash}X|>
    {\centering\arraybackslash}X|>
    {\centering\arraybackslash}X|>
    {\centering\arraybackslash}X|>
    {\centering\arraybackslash} X|}
    
        \hline 
        \textbf{Methods} & \textbf{Game} & \textbf{Gest.} & \textbf{Greeting} & \textbf{Lower Stretch} & \textbf{Pat} & \textbf{React} & \textbf{Talk} & \textbf{Upper Stretch} & 
        \textbf{Walk} & \textbf{All}\\
        \hline
       xR-EgoPose \cite{xregopose} & 56.0 & 50.2 & 44.6 & 51.5 & 59.4 & 60.8 & 43.9 & 53.9 & 57.7 & 58.2\\ \hline
       SelfPose \cite{selfpose} & 60.4 & 54.6 & 44.7 & 56.5 & 57.7 & 52.7 & 56.4 & 53.6 & 55.4 & 54.7\\ \hline
       Zhang et al. \cite{autocalib} & 36.8 & 34.1 & 36.7 & 50.1 & 57.2 & 34.4 & 32.8 & 54.3 & 52.6 & 50.0\\ \hline
       EgoFish3D \cite{egofish3d} & 48.0 & 48.2 & 42.5 & 47.3 & 48.8 & 53.6 & 47.2 & 36.2 & 48.9 & 46.1\\ \hline
       Ego-STAN \cite{egostan} & 33.1 & 31.6 & 36.9 & 38.9 & 29.2 & 29.6 & 29.7 & 44.3 & 40.9 & 40.4\\ \hline
       \textbf{EgoGlass} \cite{egoglass} & 32.8 & 30.5 & 33.7 & 35.5 & 45.7 & 33.2 & 27.0 & 40.1 & 37.4 & \textbf{37.7}\\ \hline
    \end{tabularx}
  \vspace{-3mm}
  \caption{Comparison of different skeletal based 3D egocentric pose estimation methods on xR-EgoPose dataset using MPJPE (mm).}
   \label{tab:xr-egopose}
\end{table*}

Dittadi et al. \cite{partial} used variational autoencoders to generate human body poses from \textbf{noisy head and hand pose data}. It addresses the challenge of predicting full body poses with limited information by training specialized inference models.
Yuan et al. \cite{imitation} employed control-based modeling with physics simulation and used imitation learning to acquire a video-conditioned control policy for ego-pose estimation. Traditional computer vision methods focus solely on motion kinematics \cite{kinematics} neglecting the underlying physics of dynamics \cite{dynamics}. Taking this into account, this framework allows domain adaptation, transferring the policy from simulation to real-world data.
\textit{CoolMoves} \cite{coolmoves} is a VR system that has achieved real-time, expressive full-body motion synthesis for a user's avatar using limited input cues from VR systems. It delineates the prominent movements through dynamic fusion with stylized examples from skilled performers. The system excels in synthesizing upper and lower-body motions without explicit tracking cues, addressing challenges in activities like dancing and fighting.

To solve the problem of \textbf{top-down view distortions}, \textit{EgoRenderer} \cite{egorenderer} renderes full-body avatars from egocentric images by decomposing the process into texture synthesis, pose construction, and neural image translation \cite{neuralrendering,neuralrendering2}.
\textit{The Human POSEitioning System (HPS)} \cite{hps} combines wearable sensors, IMUs, and a head-mounted camera to accurately track and integrate 3D human poses within pre-scanned environments. By fusing camera-based localization with IMU-based tracking and scene constraints, HPS achieves physically plausible motion estimation.

To address challenges like \textbf{body truncation} and \textbf{pose ambiguities}, \cite{egohmr} introduced a scene-conditioned diffusion model guided by a physics-based collision score, facilitating the generation of realistic human-scene interactions. It uses classifier-free training for flexibility in sampling, providing accurate estimations for visible body parts and diverse plausible results for unseen parts. \cite{avatarposer} predicted full-body poses in world coordinates solely from motion input derived from the user's head and hands. Leveraging a transformer encoder, the method extracts deep features, distinguishing global motion from local joint orientations to facilitate pose estimation. 
\textit{FLAG} \cite{flag} uses sparse input signals from head mounted devices and a flow-based generative model to predict full-body poses and provide uncertainty estimates for joints. 
\cite{proprioception} estimated 3D wearer poses from egocentric video, overcoming challenges of \textbf{body invisibility and complex motion}. They convert raw video to 3D pose, incorporating Self-Perception Excitation module for self understanding from egocentric view.

\textit{EgoEgo} \cite{egoego} uses monocular egocentric videos to estimate ego-head pose and generate ego-body pose, allowing \textbf{independent learning without paired datasets}. It combines monocular SLAM and transformer-based models for accurate ego-head pose estimation, employing a conditional diffusion model for full-body pose generation based on the predicted head pose.
\textit{SimpleEgo} \cite{simpleego} performs regression of probabilistic full-body pose parameters from head-mounted camera images. It directly predicts joint rotations, eliminating the need for iterative fitting processes or manual tuning. By representing joint rotations as matrix Fisher distributions, the model predicts confidence scores, allowing for robust uncertainty estimation. 
\textit{AGRoL} \cite{agrol} proposes a lightweight MLP-based diffusion model for realistic full-body motion synthesis from sparse tracking signals. \cite{mocapeverywhere} introduces affordable motion capture using smartwatches and head-mounted camera, integrating head poses for sparsity, tracking floor levels for outdoor settings, and optimizing motion with visual cues.
\textit{EgoPoser} \cite{egoposer} generates full-body pose estimation using sparse motion sensors, focusing on HMD-based ego-body pose estimation in large scenes. It addresses overfitting issues by emphasizing position-invariant prediction with a Global-in-Local motion decomposition strategy. Notably, it adapts to diverse body sizes and remains robust when hands are out of view.

\section{Evaluation Metrics}
\label{sec:6_evaluation_metrics}
In this section, we briefly describe the different metrics used to assess 3D egocentric human pose estimation methods.

\textbf{\textit{MPJPE (Mean Per Joint Position Error)}} is a widely utilized metric which measures the mean error between all the predicted 3D joint positions and the ground truth positions, by calculating the Euclidean distance between them.


\textbf{\textit{PA-MPJPE}} focuses on the individual pose accuracy by checking the alignment between the estimated pose and the ground truth pose of each frame using Procrustes analysis.

\textbf{\textit{BA-MPJPE}} first resizes the bone lengths to a standard skeleton and then calculates the PA-MPJPE, providing a comprehensive evaluation by considering structural consistency in bone lengths and eliminating body scale influence.

\textbf{\textit{Global MPJPE}} evaluates global joint position accuracy by aligning all poses within a batch to the ground truth, considering translation and rotation.

\textbf{\textit{MPJRE (Mean Per Joint Rotation Error)}} and \textbf{\textit{MPJVE (Mean Per Joint Velocity Error)}} compares the predicted and ground truth joints by calculating the average rotational and velocity disparity respectively.

\textbf{\textit{Percentage of Correct Key-points (PCK)}} is a measure of accuracy that checks if the predicted keypoint and the actual joint are close within a specific distance limit. Typically, this distance threshold is set based on the size of the subject.

\textbf{\textit{Head Translation \& Orientation Error}} focuses on translation and rotational accuracy in head pose estimation respectively. The translation error is quantified using the mean Euclidean distance between predicted and ground truth head trajectories. Whereas, the orientation error is calculated using the Frobenius norm of the difference between the predicted and ground truth head rotation matrices.

\section{Performance Analysis}
\label{sec:8_performance_analysis}

          

In this section, we compare the performance of different state-of-the-art methods for the 3D egocentric human pose estimation on some of the popular egocentric datasets.
 
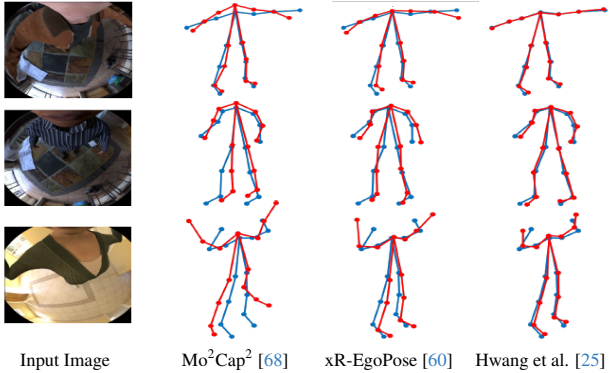
\begin{figure}[!t]
    \centering
    \small
    \begin{tikzpicture}[font=\scriptsize] 
        \node[anchor=south west, inner sep=0] (image) at (0,0) {\includegraphics[height = 45mm, width=0.995\linewidth]{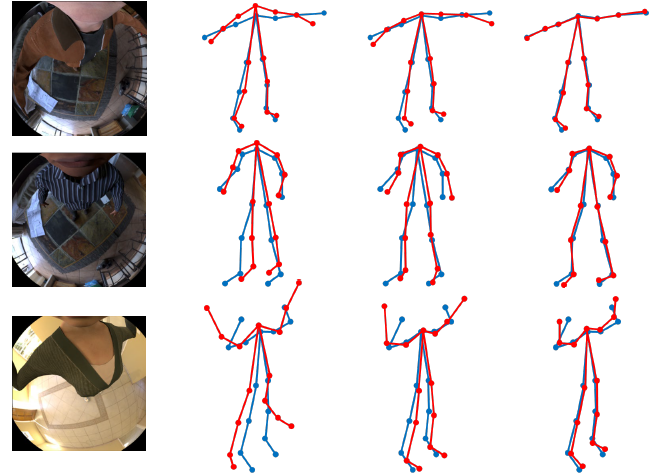}};
        \node[above right] at (0.25,-0.5) {Input Image};
        
        \node[above right] at (2.4,-0.5) {Mo\textsuperscript{2}Cap\textsuperscript{2} \cite{mo2cap2}};
        
        \node[above right] at (4.3,-0.5) {xR-EgoPose \cite{xregopose}};

        \node[above right] at (6.3,-0.5) {Hwang et al. \cite{doublediscrete}};
    \end{tikzpicture}
    \vspace{-7mm}
    \caption{Qualitative comparison between three different state-of-the-art skeletal-based egocentric 3D pose estimation models on the xR-EgoPose dataset \cite{xregopose}. The predicted 3D poses (red) are superimposed onto the ground truth poses (blue).}
    \label{fig:qualitative-result}
\vspace{-5mm}
\end{figure}

\textbf{\textit{Performance of Skeletal-based Methods: }}
Table \ref{tab:mo2cap2} shows the performance of five different skeletal-based egocentric pose estimation methods across the different actions on the widely used {Mo\textsuperscript{2}Cap\textsuperscript{2}} \cite{mo2cap2} dataset. The average MPJPE across all actions reduces from 66.8 mm in EgoFish3D \cite{egofish3d} method to
46.61 mm in SelfPose-UNet \cite{selfpose} method. We see that 2D-3D lifting models \cite{selfpose, xregopose} achieved better results than direct 3D pose estimation methods, which may be due to the preserved uncertainty information of the joints.
Figure \ref{fig:qualitative-result} shows the qualitative evaluation of the 3D poses generated by three different skeletal-based methods on the xR-EgoPose \cite{xregopose} dataset.
Table \ref{tab:xr-egopose} compares six different skeletal based methods on the xR-EgoPose \cite{xregopose} dataset. Overall, the MPJPE of the methods is lower here than those tested in {Mo\textsuperscript{2}Cap\textsuperscript{2}} \cite{mo2cap2} dataset, especially for actions with less visible joints. This could be because in this dataset most of the actions used for evaluation are relatively simpler.


\begin{table}[!b]
\vspace{-5mm}
    \centering
    \small
    \begin{tabularx}{0.46\textwidth}{|>{\centering\arraybackslash}p{3cm}|>
    {\centering\arraybackslash}p{2cm}|>
    {\centering\arraybackslash}X|}
        \hline 
        \textbf{Methods} & \textbf{MPJPE} & \textbf{MPJVE}\\ \hline
        CoolMoves \cite{coolmoves} & 7.83 & 100.54\\ \hline
        Lee et al. \cite{mocapeverywhere} & 5.87 & 19.11\\ \hline
        AvatarPoser \cite{avatarposer} & 4.18 & 29.40\\ \hline
        EgoPoser \cite{egoposer} & 4.14 & 25.95\\ \hline
        AGRoL \cite{agrol} & 3.86 & 50.94\\ \hline
        \textbf{AGRoL-Offline} \cite{agrol} & \textbf{3.71} & \textbf{18.59}\\ \hline
    \end{tabularx}
 \vspace{-2mm}
 \caption{Comparison of different body shape based 3D Egocentric Pose Estimation methods on AMASS \cite{amass} dataset using MPJPE (cm) and MPJVE (cm/s).}
 \label{tab:hpsamass}
\end{table}

\begin{figure}[!t]
    \centering
    \small
    \begin{tikzpicture}[font=\scriptsize] 
        \node[anchor=south west, inner sep=0] (image) at (0,0) {\includegraphics[height = 22mm, width=0.995\linewidth]{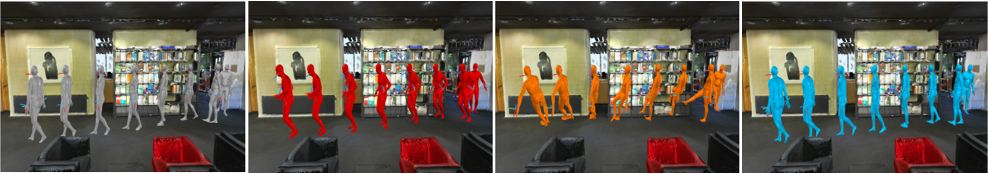}};
        \node[above right] at (0.02,-0.5) {Groundtruth Image};
        
        \node[above right] at (2.25,-0.5) {AvatarPoser \cite{avatarposer}};
        
        \node[above right] at (4.35,-0.5) {AgRoL \cite{agrol}};

        \node[above right] at (6.4,-0.5) {EgoPoser \cite{egoposer}};
    \end{tikzpicture}
    \vspace{-7mm}
    \caption{Qualitative comparison of three different state-of-the-art body shape based egocentric 3D pose estimation models on the HPS dataset. \cite{hps}.}
    \label{fig:qualitative-result-hps}
\vspace{-5mm}
\end{figure}
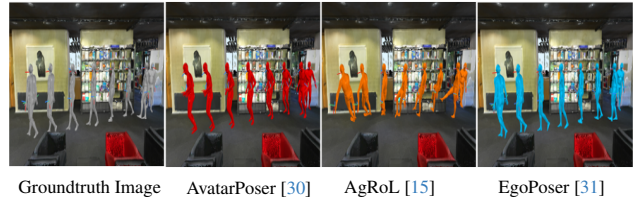

\textbf{\textit{Performance of body shape based Methods: }}
Table \ref{tab:hpsamass} shows the performance of six different body shape based egocentric pose estimation methods on the AMASS \cite{amass} dataset. We can see that, AGRoL \cite{agrol} outperforms other methods with its smooth motion generation, but it's limited to offline use. For real-time applications, EgoPoser \cite{egoposer} is more suitable as it provides more adaptability to diverse body sizes as well as robustness with hands out of view.
Figure \ref{fig:qualitative-result-hps} qualitatively compares the 3D human pose and shape on three different body shape based methods using HPS \cite{hps} dataset.



\section{Conclusion and Future Directions}
\label{sec:9_conclusion_future_direction}
In this survey paper, we provide an overview of 3D egocentric human pose estimation using RGB images or video sequences, encompassing diverse datasets and estimation methodologies. 
Researchers have proposed diverse datasets with lightweight setups. However, the lack of standardized \textbf{benchmark datasets}, except for the recent Ego-Exo4D \cite{egoexo} dataset, poses a challenge for evaluating the robustness of different egocentric pose estimation models.
While discussing the individual strengths and weaknesses of different skeletal and body shape based methods for egocentric pose estimation, we realize that most of the existing methods encounter difficulties with \textbf{in-the-wild scenarios} mainly due to insufficient training data. 
Notably, similar to traditional pose estimation, the biggest challenges of egocentric pose estimation models are strong occlusions and limited field of view, especially for the lower body joints. \textbf{Multi view consistency} may help to to solve this using additional 3D information. Moreover, \textbf{temporal and contextual information} can be utilized further to improve the robustness of the models considering these issues.
Consequently, there exists ample scope for refining egocentric pose estimation approaches to better suit \textbf{real-time} technologies.

In conclusion, this survey paper serves as a comprehensive resource for researchers seeking to explore the existing egocentric pose estimation methods, understand prevalent challenges, and make further advancements.

\section{Acknowledgements}
This material is partially based upon work supported by the National Science Foundation under Grant No. 2153249 and 2316240. Any opinions, findings, and conclusions or recommendations expressed in this material are those of the author(s) and do not necessarily reflect the views of the National Science Foundation.

\small
\bibliographystyle{cvpr24-template/ieeenat_fullname}
\bibliography{ego-pose}

\begin{thebibliography}{80}
\providecommand{\natexlab}[1]{#1}
\providecommand{\url}[1]{\texttt{#1}}
\expandafter\ifx\csname urlstyle\endcsname\relax
  \providecommand{\doi}[1]{doi: #1}\else
  \providecommand{\doi}{doi: \begingroup \urlstyle{rm}\Url}\fi

\bibitem[cmu()]{cmumocap}
{CMU Motion Capture Database}.
\newblock \url{http://mocap.cs.cmu.edu}.
\newblock Accessed: March 2, 2024.

\bibitem[Ahuja et~al.(2021)Ahuja, Ofek, Gonzalez-Franco, Holz, and Wilson]{coolmoves}
Karan Ahuja, Eyal Ofek, Mar Gonzalez-Franco, Christian Holz, and Andrew~D Wilson.
\newblock Coolmoves: User motion accentuation in virtual reality.
\newblock \emph{Proceedings of the ACM on Interactive, Mobile, Wearable and Ubiquitous Technologies}, 5\penalty0 (2):\penalty0 1--23, 2021.

\bibitem[Akada et~al.(2022)Akada, Wang, Shimada, Takahashi, Theobalt, and Golyanik]{unrealego}
Hiroyasu Akada, Jian Wang, Soshi Shimada, Masaki Takahashi, Christian Theobalt, and Vladislav Golyanik.
\newblock Unrealego: A new dataset for robust egocentric 3d human motion capture.
\newblock In \emph{European Conference on Computer Vision}, pages 1--17. Springer, 2022.

\bibitem[Akada et~al.(2023)Akada, Wang, Golyanik, and Theobalt]{unrealego2}
Hiroyasu Akada, Jian Wang, Vladislav Golyanik, and Christian Theobalt.
\newblock 3d human pose perception from egocentric stereo videos, 2023.

\bibitem[Aliakbarian et~al.(2022)Aliakbarian, Cameron, Bogo, Fitzgibbon, and Cashman]{flag}
Sadegh Aliakbarian, Pashmina Cameron, Federica Bogo, Andrew Fitzgibbon, and Thomas~J Cashman.
\newblock Flag: Flow-based 3d avatar generation from sparse observations.
\newblock In \emph{Proceedings of the IEEE/CVF Conference on Computer Vision and Pattern Recognition}, pages 13253--13262, 2022.

\bibitem[Andriluka et~al.(2014)Andriluka, Pishchulin, Gehler, and Schiele]{mpii}
Mykhaylo Andriluka, Leonid Pishchulin, Peter Gehler, and Bernt Schiele.
\newblock 2d human pose estimation: New benchmark and state of the art analysis.
\newblock In \emph{Proceedings of the IEEE Conference on computer Vision and Pattern Recognition}, pages 3686--3693, 2014.

\bibitem[Bandini and Zariffa(2020)]{handsurvey}
Andrea Bandini and Jos{\'e} Zariffa.
\newblock Analysis of the hands in egocentric vision: A survey.
\newblock \emph{IEEE transactions on pattern analysis and machine intelligence}, 2020.

\bibitem[Bogo et~al.(2016)Bogo, Kanazawa, Lassner, Gehler, Romero, and Black]{keepitsmpl}
Federica Bogo, Angjoo Kanazawa, Christoph Lassner, Peter Gehler, Javier Romero, and Michael~J Black.
\newblock Keep it smpl: Automatic estimation of 3d human pose and shape from a single image.
\newblock In \emph{Computer Vision--ECCV 2016: 14th European Conference, Amsterdam, The Netherlands, October 11-14, 2016, Proceedings, Part V 14}, pages 561--578. Springer, 2016.

\bibitem[Cao et~al.(2020)Cao, Gao, Mangalam, Cai, Vo, and Malik]{scenecontext}
Zhe Cao, Hang Gao, Karttikeya Mangalam, Qi-Zhi Cai, Minh Vo, and Jitendra Malik.
\newblock Long-term human motion prediction with scene context.
\newblock In \emph{Computer Vision--ECCV 2020: 16th European Conference, Glasgow, UK, August 23--28, 2020, Proceedings, Part I 16}, pages 387--404. Springer, 2020.

\bibitem[Cao et~al.(2021)Cao, Hidalgo, Simon, Wei, and Sheikh]{ref02}
Zhe Cao, Gines Hidalgo, Tomas Simon, Shih-En Wei, and Yaser Sheikh.
\newblock Openpose: Realtime multi-person 2d pose estimation using part affinity fields.
\newblock \emph{IEEE Transactions on Pattern Analysis and Machine Intelligence}, 2021.

\bibitem[Cuevas-Velasquez et~al.(2024)Cuevas-Velasquez, Hewitt, Aliakbarian, and Baltru{\v{s}}aitis]{simpleego}
Hanz Cuevas-Velasquez, Charlie Hewitt, Sadegh Aliakbarian, and Tadas Baltru{\v{s}}aitis.
\newblock Simpleego: Predicting probabilistic body pose from egocentric cameras.
\newblock \emph{arXiv preprint arXiv:2401.14785}, 2024.

\bibitem[Dhamanaskar et~al.(2023{\natexlab{a}})Dhamanaskar, Dimiccoli, Corona, Pumarola, and Moreno-Noguer]{dhamanaskar}
Ameya Dhamanaskar, Mariella Dimiccoli, Enric Corona, Albert Pumarola, and Francesc Moreno-Noguer.
\newblock Enhancing egocentric 3d pose estimation with third person views.
\newblock \emph{Pattern Recognition}, 138:\penalty0 109358, 2023{\natexlab{a}}.

\bibitem[Dhamanaskar et~al.(2023{\natexlab{b}})Dhamanaskar, Dimiccoli, Corona, Pumarola, and Moreno-Noguer]{first2third}
Ameya Dhamanaskar, Mariella Dimiccoli, Enric Corona, Albert Pumarola, and Francesc Moreno-Noguer.
\newblock Enhancing egocentric 3d pose estimation with third person views.
\newblock \emph{Pattern Recognition}, 138:\penalty0 109358, 2023{\natexlab{b}}.

\bibitem[Dittadi et~al.(2021)Dittadi, Dziadzio, Cosker, Lundell, Cashman, and Shotton]{partial}
Andrea Dittadi, Sebastian Dziadzio, Darren Cosker, Ben Lundell, Thomas~J Cashman, and Jamie Shotton.
\newblock Full-body motion from a single head-mounted device: Generating smpl poses from partial observations.
\newblock In \emph{Proceedings of the IEEE/CVF International Conference on Computer Vision}, pages 11687--11697, 2021.

\bibitem[Du et~al.(2023)Du, Kips, Pumarola, Starke, Thabet, and Sanakoyeu]{agrol}
Yuming Du, Robin Kips, Albert Pumarola, Sebastian Starke, Ali Thabet, and Artsiom Sanakoyeu.
\newblock Avatars grow legs: Generating smooth human motion from sparse tracking inputs with diffusion model.
\newblock In \emph{Proceedings of the IEEE/CVF Conference on Computer Vision and Pattern Recognition (CVPR)}, pages 481--490, 2023.

\bibitem[Gamra and Akhloufi(2021)]{ref09}
Miniar~Ben Gamra and Moulay~A Akhloufi.
\newblock A review of deep learning techniques for 2d and 3d human pose estimation.
\newblock \emph{Image and Vision Computing}, 114:\penalty0 104282, 2021.

\bibitem[Garrido-Jurado et~al.(2014)Garrido-Jurado, Mu{\~n}oz-Salinas, Madrid-Cuevas, and Mar{\'\i}n-Jim{\'e}nez]{arucomarker}
Sergio Garrido-Jurado, Rafael Mu{\~n}oz-Salinas, Francisco~Jos{\'e} Madrid-Cuevas, and Manuel~Jes{\'u}s Mar{\'\i}n-Jim{\'e}nez.
\newblock Automatic generation and detection of highly reliable fiducial markers under occlusion.
\newblock \emph{Pattern Recognition}, 47\penalty0 (6):\penalty0 2280--2292, 2014.

\bibitem[Grauman et~al.(2023)Grauman, Westbury, Torresani, Kitani, Malik, Afouras, Ashutosh, Baiyya, Bansal, Boote, et~al.]{egoexo}
Kristen Grauman, Andrew Westbury, Lorenzo Torresani, Kris Kitani, Jitendra Malik, Triantafyllos Afouras, Kumar Ashutosh, Vijay Baiyya, Siddhant Bansal, Bikram Boote, et~al.
\newblock Ego-exo4d: Understanding skilled human activity from first-and third-person perspectives.
\newblock \emph{arXiv preprint arXiv:2311.18259}, 2023.

\bibitem[G{\"u}ler et~al.(2018)G{\"u}ler, Neverova, and Kokkinos]{ref04}
R{\i}za~Alp G{\"u}ler, Natalia Neverova, and Iasonas Kokkinos.
\newblock Densepose: Dense human pose estimation in the wild.
\newblock In \emph{Proceedings of the IEEE conference on computer vision and pattern recognition}, pages 7297--7306, 2018.

\bibitem[Guzov et~al.(2021)Guzov, Mir, Sattler, and Pons-Moll]{hps}
Vladimir Guzov, Aymen Mir, Torsten Sattler, and Gerard Pons-Moll.
\newblock Human poseitioning system (hps): 3d human pose estimation and self-localization in large scenes from body-mounted sensors.
\newblock In \emph{Proceedings of the IEEE/CVF Conference on Computer Vision and Pattern Recognition}, pages 4318--4329, 2021.

\bibitem[Hu et~al.(2021)Hu, Sarkar, Liu, Zwicker, and Theobalt]{egorenderer}
Tao Hu, Kripasindhu Sarkar, Lingjie Liu, Matthias Zwicker, and Christian Theobalt.
\newblock Egorenderer: Rendering human avatars from egocentric camera images.
\newblock In \emph{Proceedings of the IEEE/CVF International Conference on Computer Vision}, pages 14528--14538, 2021.

\bibitem[Hu et~al.(2022)Hu, Yu, Zheng, Zhang, Liu, and Zwicker]{rendering}
Tao Hu, Tao Yu, Zerong Zheng, He Zhang, Yebin Liu, and Matthias Zwicker.
\newblock Hvtr: Hybrid volumetric-textural rendering for human avatars.
\newblock In \emph{2022 International Conference on 3D Vision (3DV)}, pages 197--208. IEEE, 2022.

\bibitem[Huang et~al.(2017)Huang, Bogo, Lassner, Kanazawa, Gehler, Romero, Akhter, and Black]{towardsaccurate}
Yinghao Huang, Federica Bogo, Christoph Lassner, Angjoo Kanazawa, Peter~V Gehler, Javier Romero, Ijaz Akhter, and Michael~J Black.
\newblock Towards accurate marker-less human shape and pose estimation over time.
\newblock In \emph{2017 international conference on 3D vision (3DV)}, pages 421--430. IEEE, 2017.

\bibitem[Huang et~al.(2020)Huang, Xu, Lassner, Li, and Tung]{arch}
Zeng Huang, Yuanlu Xu, Christoph Lassner, Hao Li, and Tony Tung.
\newblock Arch: Animatable reconstruction of clothed humans.
\newblock In \emph{Proceedings of the IEEE/CVF Conference on Computer Vision and Pattern Recognition}, pages 3093--3102, 2020.

\bibitem[Hwang and Kang(2024)]{doublediscrete}
Juheon Hwang and Jiwoo Kang.
\newblock Double discrete representation for 3d human pose estimation from head-mounted camera.
\newblock In \emph{2024 IEEE International Conference on Consumer Electronics (ICCE)}, pages 1--4. IEEE, 2024.

\bibitem[Ionescu et~al.(2013)Ionescu, Papava, Olaru, and Sminchisescu]{human3.6m}
Catalin Ionescu, Dragos Papava, Vlad Olaru, and Cristian Sminchisescu.
\newblock Human3. 6m: Large scale datasets and predictive methods for 3d human sensing in natural environments.
\newblock \emph{IEEE transactions on pattern analysis and machine intelligence}, 36\penalty0 (7):\penalty0 1325--1339, 2013.

\bibitem[Ji et~al.(2020)Ji, Fang, Dong, Shuai, Jiang, and Zhou]{ref05}
Xiaopeng Ji, Qi Fang, Junting Dong, Qing Shuai, Wen Jiang, and Xiaowei Zhou.
\newblock A survey on monocular 3d human pose estimation.
\newblock \emph{Virtual Reality \& Intelligent Hardware}, 2:\penalty0 471--500, 2020.

\bibitem[Jiang and Grauman(2017)]{invisible}
Hao Jiang and Kristen Grauman.
\newblock Seeing invisible poses: Estimating 3d body pose from egocentric video.
\newblock In \emph{2017 IEEE Conference on Computer Vision and Pattern Recognition (CVPR)}, pages 3501--3509. IEEE, 2017.

\bibitem[Jiang and Ithapu(2021)]{visionspan}
Hao Jiang and Vamsi~Krishna Ithapu.
\newblock Egocentric pose estimation from human vision span.
\newblock In \emph{2021 IEEE/CVF International Conference on Computer Vision (ICCV)}, pages 10986--10994. IEEE, 2021.

\bibitem[Jiang et~al.(2022)Jiang, Streli, Qiu, Fender, Laich, Snape, and Holz]{avatarposer}
Jiaxi Jiang, Paul Streli, Huajian Qiu, Andreas Fender, Larissa Laich, Patrick Snape, and Christian Holz.
\newblock Avatarposer: Articulated full-body pose tracking from sparse motion sensing.
\newblock In \emph{European Conference on Computer Vision}, pages 443--460. Springer, 2022.

\bibitem[Jiang et~al.(2023)Jiang, Streli, Meier, and Holz]{egoposer}
Jiaxi Jiang, Paul Streli, Manuel Meier, and Christian Holz.
\newblock Egoposer: Robust real-time ego-body pose estimation in large scenes, 2023.

\bibitem[Kang and Lee(2024)]{egotap}
Taeho Kang and Youngki Lee.
\newblock Attention-propagation network for egocentric heatmap to 3d pose lifting.
\newblock \emph{arXiv preprint arXiv:2402.18330}, 2024.

\bibitem[Kang et~al.(2023)Kang, Lee, Zhang, and Lee]{ego3dpose}
Taeho Kang, Kyungjin Lee, Jinrui Zhang, and Youngki Lee.
\newblock Ego3dpose: Capturing 3d cues from binocular egocentric views.
\newblock In \emph{SIGGRAPH Asia 2023 Conference Papers}, pages 1--10, 2023.

\bibitem[Lee and Joo(2024)]{mocapeverywhere}
Jiye Lee and Hanbyul Joo.
\newblock Mocap everyone everywhere: Lightweight motion capture with smartwatches and a head-mounted camera.
\newblock \emph{arXiv preprint arXiv:2401.00847}, 2024.

\bibitem[Li et~al.(2023{\natexlab{a}})Li, Liu, and Wu]{egoego}
Jiaman Li, Karen Liu, and Jiajun Wu.
\newblock Ego-body pose estimation via ego-head pose estimation.
\newblock In \emph{Proceedings of the IEEE/CVF Conference on Computer Vision and Pattern Recognition}, pages 17142--17151, 2023{\natexlab{a}}.

\bibitem[Li et~al.(2023{\natexlab{b}})Li, Zhang, Su, and Liu]{egoformer}
Tianyi Li, Chi Zhang, Wei Su, and Yuehu Liu.
\newblock Egoformer: Transformer-based motion context learning for ego-pose estimation.
\newblock In \emph{2023 IEEE International Conference on Systems, Man, and Cybernetics (SMC)}, pages 4052--4057. IEEE, 2023{\natexlab{b}}.

\bibitem[Li et~al.(2021)Li, Nagarajan, Xiong, and Grauman]{ego-exo}
Yanghao Li, Tushar Nagarajan, Bo Xiong, and Kristen Grauman.
\newblock Ego-exo: Transferring visual representations from third-person to first-person videos.
\newblock In \emph{Proceedings of the IEEE/CVF Conference on Computer Vision and Pattern Recognition}, pages 6943--6953, 2021.

\bibitem[Liu et~al.(2022)Liu, Yang, Gu, Guo, and Yang]{ego+x}
Yuxuan Liu, Jianxin Yang, Xiao Gu, Yao Guo, and Guang-Zhong Yang.
\newblock Ego+ x: An egocentric vision system for global 3d human pose estimation and social interaction characterization.
\newblock In \emph{2022 IEEE/RSJ International Conference on Intelligent Robots and Systems (IROS)}, pages 5271--5277. IEEE, 2022.

\bibitem[Liu et~al.(2023)Liu, Yang, Gu, Chen, Guo, and Yang]{egofish3d}
Yuxuan Liu, Jianxin Yang, Xiao Gu, Yijun Chen, Yao Guo, and Guang-Zhong Yang.
\newblock Egofish3d: Egocentric 3d pose estimation from a fisheye camera via self-supervised learning.
\newblock \emph{IEEE Transactions on Multimedia}, 25:\penalty0 8880--8891, 2023.

\bibitem[Liu et~al.(2024)Liu, Qiu, and Zhang]{meshsurvey1}
Yang Liu, Changzhen Qiu, and Zhiyong Zhang.
\newblock Deep learning for 3d human pose estimation and mesh recovery: A survey.
\newblock \emph{arXiv preprint arXiv:2402.18844}, 2024.

\bibitem[Loper et~al.(2023)Loper, Mahmood, Romero, Pons-Moll, and Black]{smpl}
Matthew Loper, Naureen Mahmood, Javier Romero, Gerard Pons-Moll, and Michael~J Black.
\newblock Smpl: A skinned multi-person linear model.
\newblock In \emph{Seminal Graphics Papers: Pushing the Boundaries, Volume 2}, pages 851--866. 2023.

\bibitem[Luo et~al.(2020)Luo, Hachiuma, Yuan, Iwase, and Kitani]{kinematics}
Zhengyi Luo, Ryo Hachiuma, Ye Yuan, Shun Iwase, and Kris~M Kitani.
\newblock Kinematics-guided reinforcement learning for object-aware 3d ego-pose estimation.
\newblock \emph{arXiv preprint arXiv:2011.04837}, 2020.

\bibitem[Luo et~al.(2021)Luo, Hachiuma, Yuan, and Kitani]{dynamics}
Zhengyi Luo, Ryo Hachiuma, Ye Yuan, and Kris Kitani.
\newblock Dynamics-regulated kinematic policy for egocentric pose estimation.
\newblock \emph{Advances in Neural Information Processing Systems}, 34:\penalty0 25019--25032, 2021.

\bibitem[Mahmood et~al.(2019)Mahmood, Ghorbani, Troje, Pons-Moll, and Black]{amass}
Naureen Mahmood, Nima Ghorbani, Nikolaus~F Troje, Gerard Pons-Moll, and Michael~J Black.
\newblock Amass: Archive of motion capture as surface shapes.
\newblock In \emph{Proceedings of the IEEE/CVF international conference on computer vision}, pages 5442--5451, 2019.

\bibitem[Munea et~al.(2020)Munea, Jembre, Weldegebriel, Chen, Huang, and Yang]{ref08}
Tewodros~Legesse Munea, Yalew~Zelalem Jembre, Halefom~Tekle Weldegebriel, Longbiao Chen, Chenxi Huang, and Chenhui Yang.
\newblock The progress of human pose estimation: A survey and taxonomy of models applied in 2d human pose estimation.
\newblock \emph{IEEE Access}, 8:\penalty0 133330--133348, 2020.

\bibitem[Ng et~al.(2020)Ng, Xiang, Joo, and Grauman]{you2me}
Evonne Ng, Donglai Xiang, Hanbyul Joo, and Kristen Grauman.
\newblock You2me: Inferring body pose in egocentric video via first and second person interactions.
\newblock In \emph{Proceedings of the IEEE/CVF Conference on Computer Vision and Pattern Recognition}, pages 9890--9900, 2020.

\bibitem[N{\'u}{\~n}ez-Marcos et~al.(2022)N{\'u}{\~n}ez-Marcos, Azkune, and Arganda-Carreras]{actionsurveyegocentric}
Adri{\'a}n N{\'u}{\~n}ez-Marcos, Gorka Azkune, and Ignacio Arganda-Carreras.
\newblock Egocentric vision-based action recognition: A survey.
\newblock \emph{Neurocomputing}, 472:\penalty0 175--197, 2022.

\bibitem[Park et~al.(2023)Park, Kaai, Hossain, Sumi, Rambhatla, and Fieguth]{egostan}
Jinman Park, Kimathi Kaai, Saad Hossain, Norikatsu Sumi, Sirisha Rambhatla, and Paul Fieguth.
\newblock Domain-guided spatio-temporal self-attention for egocentric 3d pose estimation.
\newblock In \emph{Proceedings of the 29th ACM SIGKDD Conference on Knowledge Discovery and Data Mining}, pages 1837--1849, 2023.

\bibitem[Pavlakos et~al.(2018)Pavlakos, Zhou, and Daniilidis]{skeletal1}
Georgios Pavlakos, Xiaowei Zhou, and Kostas Daniilidis.
\newblock Ordinal depth supervision for 3d human pose estimation.
\newblock In \emph{Proceedings of the IEEE conference on computer vision and pattern recognition}, pages 7307--7316, 2018.

\bibitem[Pavlakos et~al.(2019)Pavlakos, Choutas, Ghorbani, Bolkart, Osman, Tzionas, and Black]{smplx}
Georgios Pavlakos, Vasileios Choutas, Nima Ghorbani, Timo Bolkart, Ahmed~AA Osman, Dimitrios Tzionas, and Michael~J Black.
\newblock Expressive body capture: 3d hands, face, and body from a single image.
\newblock In \emph{Proceedings of the IEEE/CVF conference on computer vision and pattern recognition}, pages 10975--10985, 2019.

\bibitem[Rhodin et~al.(2016)Rhodin, Richardt, Casas, Insafutdinov, Shafiei, Seidel, Schiele, and Theobalt]{egocap}
Helge Rhodin, Christian Richardt, Dan Casas, Eldar Insafutdinov, Mohammad Shafiei, Hans-Peter Seidel, Bernt Schiele, and Christian Theobalt.
\newblock Egocap: egocentric marker-less motion capture with two fisheye cameras.
\newblock \emph{ACM Transactions on Graphics (TOG)}, 35\penalty0 (6):\penalty0 1--11, 2016.

\bibitem[Sitzmann et~al.(2019)Sitzmann, Zollh{\"o}fer, and Wetzstein]{neuralrendering}
Vincent Sitzmann, Michael Zollh{\"o}fer, and Gordon Wetzstein.
\newblock Scene representation networks: Continuous 3d-structure-aware neural scene representations.
\newblock \emph{Advances in Neural Information Processing Systems}, 32, 2019.

\bibitem[Soran et~al.(2015)Soran, Farhadi, and Shapiro]{third-person1}
Bilge Soran, Ali Farhadi, and Linda Shapiro.
\newblock Action recognition in the presence of one egocentric and multiple static cameras.
\newblock In \emph{Computer Vision--ACCV 2014: 12th Asian Conference on Computer Vision, Singapore, Singapore, November 1-5, 2014, Revised Selected Papers, Part V 12}, pages 178--193. Springer, 2015.

\bibitem[Su et~al.(2022)Su, Liu, Li, and Cai]{proprioception}
Wei Su, Yuehu Liu, Shasha Li, and Zerun Cai.
\newblock Proprioception-driven wearer pose estimation for egocentric video.
\newblock In \emph{2022 26th International Conference on Pattern Recognition (ICPR)}, pages 3728--3735. IEEE, 2022.

\bibitem[Sun et~al.(2019)Sun, Xiao, Liu, and Wang]{ref03}
Ke Sun, Bin Xiao, Dong Liu, and Jingdong Wang.
\newblock Deep high-resolution representation learning for human pose estimation.
\newblock In \emph{Proceedings of the IEEE/CVF conference on computer vision and pattern recognition}, 2019.

\bibitem[Sun et~al.(2017)Sun, Shang, Liang, and Wei]{skeletal2}
Xiao Sun, Jiaxiang Shang, Shuang Liang, and Yichen Wei.
\newblock Compositional human pose regression.
\newblock In \emph{Proceedings of the IEEE international conference on computer vision}, pages 2602--2611, 2017.

\bibitem[Tekin et~al.(2016)Tekin, Katircioglu, Salzmann, Lepetit, and Fua]{skeletal3}
Bugra Tekin, Isinsu Katircioglu, Mathieu Salzmann, Vincent Lepetit, and Pascal Fua.
\newblock Structured prediction of 3d human pose with deep neural networks.
\newblock \emph{arXiv preprint arXiv:1605.05180}, 2016.

\bibitem[Thies et~al.(2019)Thies, Zollh{\"o}fer, and Nie{\ss}ner]{neuralrendering2}
Justus Thies, Michael Zollh{\"o}fer, and Matthias Nie{\ss}ner.
\newblock Deferred neural rendering: Image synthesis using neural textures.
\newblock \emph{Acm Transactions on Graphics (TOG)}, 38\penalty0 (4):\penalty0 1--12, 2019.

\bibitem[Tian et~al.(2023)Tian, Zhang, Liu, and Wang]{meshsurvey2}
Yating Tian, Hongwen Zhang, Yebin Liu, and Limin Wang.
\newblock Recovering 3d human mesh from monocular images: A survey.
\newblock \emph{IEEE transactions on pattern analysis and machine intelligence}, 2023.

\bibitem[Tome et~al.(2019)Tome, Peluse, Agapito, and Badino]{xregopose}
Denis Tome, Patrick Peluse, Lourdes Agapito, and Hernan Badino.
\newblock xr-egopose: Egocentric 3d human pose from an hmd camera.
\newblock In \emph{Proceedings of the IEEE/CVF International Conference on Computer Vision}, pages 7728--7738, 2019.

\bibitem[Tome et~al.(2023)Tome, Alldieck, Peluse, Pons-Moll, Agapito, Badino, and de~la Torre]{selfpose}
Denis Tome, Thiemo Alldieck, Patrick Peluse, Gerard Pons-Moll, Lourdes Agapito, Hernan Badino, and Fernando de~la Torre.
\newblock Selfpose: 3d egocentric pose estimation from a headset mounted camera.
\newblock \emph{IEEE Transactions on Pattern Analysis and Machine Intelligence}, 45\penalty0 (6):\penalty0 6794--6806, 2023.

\bibitem[Toshev and Szegedy(2014)]{ref01}
Alexander Toshev and Christian Szegedy.
\newblock Deeppose: Human pose estimation via deep neural networks.
\newblock In \emph{2014 IEEE Conference on Computer Vision and Pattern Recognition}, 2014.

\bibitem[Varol et~al.(2017)Varol, Romero, Martin, Mahmood, Black, Laptev, and Schmid]{surreal}
Gul Varol, Javier Romero, Xavier Martin, Naureen Mahmood, Michael~J Black, Ivan Laptev, and Cordelia Schmid.
\newblock Learning from synthetic humans.
\newblock In \emph{Proceedings of the IEEE conference on computer vision and pattern recognition}, pages 109--117, 2017.

\bibitem[Wang et~al.(2021{\natexlab{a}})Wang, Liu, Xu, Sarkar, and Theobalt]{egoglobal}
Jian Wang, Lingjie Liu, Weipeng Xu, Kripasindhu Sarkar, and Christian Theobalt.
\newblock Estimating egocentric 3d human pose in global space.
\newblock In \emph{Proceedings of the IEEE/CVF International Conference on Computer Vision}, pages 11500--11509, 2021{\natexlab{a}}.

\bibitem[Wang et~al.(2021{\natexlab{b}})Wang, Tan, Zhen, Xu, Zheng, He, and Shao]{ref07}
Jinbao Wang, Shujie Tan, Xiantong Zhen, Shuo Xu, Feng Zheng, Zhenyu He, and Ling Shao.
\newblock Deep 3d human pose estimation: A review.
\newblock \emph{Computer Vision and Image Understanding}, 210:\penalty0 103225, 2021{\natexlab{b}}.

\bibitem[Wang et~al.(2022)Wang, Liu, Xu, Sarkar, Luvizon, and Theobalt]{egopw}
Jian Wang, Lingjie Liu, Weipeng Xu, Kripasindhu Sarkar, Diogo Luvizon, and Christian Theobalt.
\newblock Estimating egocentric 3d human pose in the wild with external weak supervision.
\newblock In \emph{Proceedings of the IEEE/CVF Conference on Computer Vision and Pattern Recognition}, pages 13157--13166, 2022.

\bibitem[Wang et~al.(2023)Wang, Luvizon, Xu, Liu, Sarkar, and Theobalt]{sceneaware}
Jian Wang, Diogo Luvizon, Weipeng Xu, Lingjie Liu, Kripasindhu Sarkar, and Christian Theobalt.
\newblock Scene-aware egocentric 3d human pose estimation.
\newblock In \emph{Proceedings of the IEEE/CVF Conference on Computer Vision and Pattern Recognition}, pages 13031--13040, 2023.

\bibitem[Xu et~al.(2019)Xu, Chatterjee, Zollhoefer, Rhodin, Fua, Seidel, and Theobalt]{mo2cap2}
Weipeng Xu, Avishek Chatterjee, Michael Zollhoefer, Helge Rhodin, Pascal Fua, Hans-Peter Seidel, and Christian Theobalt.
\newblock Mo 2 cap 2: Real-time mobile 3d motion capture with a cap-mounted fisheye camera.
\newblock \emph{IEEE transactions on visualization and computer graphics}, 25\penalty0 (5):\penalty0 2093--2101, 2019.

\bibitem[Yonetani et~al.(2016)Yonetani, Kitani, and Sato]{third-person2}
Ryo Yonetani, Kris~M Kitani, and Yoichi Sato.
\newblock Recognizing micro-actions and reactions from paired egocentric videos.
\newblock In \emph{Proceedings of the IEEE Conference on Computer Vision and Pattern Recognition}, pages 2629--2638, 2016.

\bibitem[Yu et~al.(2018)Yu, Zheng, Guo, Zhao, Dai, Li, Pons-Moll, and Liu]{clothed1}
Tao Yu, Zerong Zheng, Kaiwen Guo, Jianhui Zhao, Qionghai Dai, Hao Li, Gerard Pons-Moll, and Yebin Liu.
\newblock Doublefusion: Real-time capture of human performances with inner body shapes from a single depth sensor.
\newblock In \emph{Proceedings of the IEEE conference on computer vision and pattern recognition}, pages 7287--7296, 2018.

\bibitem[Yuan and Kitani(2018)]{imitation}
Ye Yuan and Kris Kitani.
\newblock 3d ego-pose estimation via imitation learning.
\newblock In \emph{Proceedings of the European Conference on Computer Vision (ECCV)}, pages 735--750, 2018.

\bibitem[Yuan and Kitani(2019)]{forecast}
Ye Yuan and Kris Kitani.
\newblock Ego-pose estimation and forecasting as real-time pd control.
\newblock In \emph{Proceedings of the IEEE/CVF International Conference on Computer Vision}, pages 10082--10092, 2019.

\bibitem[Zanfir et~al.(2018)Zanfir, Marinoiu, and Sminchisescu]{monocularshape}
Andrei Zanfir, Elisabeta Marinoiu, and Cristian Sminchisescu.
\newblock Monocular 3d pose and shape estimation of multiple people in natural scenes-the importance of multiple scene constraints.
\newblock In \emph{Proceedings of the IEEE conference on computer vision and pattern recognition}, pages 2148--2157, 2018.

\bibitem[Zhang et~al.(2022)Zhang, Ma, Zhang, Qian, Kwon, Pollefeys, Bogo, and Tang]{egobody}
Siwei Zhang, Qianli Ma, Yan Zhang, Zhiyin Qian, Taein Kwon, Marc Pollefeys, Federica Bogo, and Siyu Tang.
\newblock Egobody: Human body shape and motion of interacting people from head-mounted devices.
\newblock In \emph{European Conference on Computer Vision}, pages 180--200. Springer, 2022.

\bibitem[Zhang et~al.(2023)Zhang, Ma, Zhang, Aliakbarian, Cosker, and Tang]{egohmr}
Siwei Zhang, Qianli Ma, Yan Zhang, Sadegh Aliakbarian, Darren Cosker, and Siyu Tang.
\newblock Probabilistic human mesh recovery in 3d scenes from egocentric views.
\newblock \emph{arXiv preprint arXiv:2304.06024}, 2023.

\bibitem[Zhang et~al.(2021)Zhang, You, and Gevers]{autocalib}
Yahui Zhang, Shaodi You, and Theo Gevers.
\newblock Automatic calibration of the fisheye camera for egocentric 3d human pose estimation from a single image.
\newblock In \emph{Proceedings of the IEEE/CVF Winter Conference on Applications of Computer Vision}, pages 1772--1781, 2021.

\bibitem[Zhao et~al.(2021)Zhao, Wei, Mahmud, and Frahm]{egoglass}
Dongxu Zhao, Zhen Wei, Jisan Mahmud, and Jan-Michael Frahm.
\newblock Egoglass: Egocentric-view human pose estimation from an eyeglass frame.
\newblock In \emph{2021 International Conference on 3D Vision (3DV)}, pages 32--41. IEEE, 2021.

\bibitem[Zheng et~al.(2023{\natexlab{a}})Zheng, Wu, Chen, Yang, Zhu, Shen, Kehtarnavaz, and Shah]{ref06}
Ce Zheng, Wenhan Wu, Chen Chen, Taojiannan Yang, Sijie Zhu, Ju Shen, Nasser Kehtarnavaz, and Mubarak Shah.
\newblock Deep learning-based human pose estimation: A survey.
\newblock \emph{ACM Computing Surveys}, 56\penalty0 (1):\penalty0 1--37, 2023{\natexlab{a}}.

\bibitem[Zheng et~al.(2021)Zheng, Yu, Liu, and Dai]{clothed2}
Zerong Zheng, Tao Yu, Yebin Liu, and Qionghai Dai.
\newblock Pamir: Parametric model-conditioned implicit representation for image-based human reconstruction.
\newblock \emph{IEEE transactions on pattern analysis and machine intelligence}, 44\penalty0 (6):\penalty0 3170--3184, 2021.

\bibitem[Zheng et~al.(2023{\natexlab{b}})Zheng, Zhao, Zhang, Liu, and Liu]{avatarrex}
Zerong Zheng, Xiaochen Zhao, Hongwen Zhang, Boning Liu, and Yebin Liu.
\newblock Avatarrex: Real-time expressive full-body avatars.
\newblock \emph{ACM Transactions on Graphics (TOG)}, 42\penalty0 (4):\penalty0 1--19, 2023{\natexlab{b}}.

\end{thebibliography}


\end{document}